\documentclass[sigconf]{acmart}
\usepackage{amsmath}
\usepackage{amsthm}
\usepackage{algorithm}
\usepackage{algorithmic}
\usepackage{multirow}
\usepackage{balance}

\AtBeginDocument{%
  }

\copyrightyear{2023}
\acmYear{2023}
\setcopyright{acmlicensed}
\acmConference[MM '23] {Proceedings of the 31st ACM International Conference on Multimedia}{October 29--November 3, 2023}{Ottawa, ON, Canada.}
\acmBooktitle{Proceedings of the 31st ACM International Conference on Multimedia (MM '23), October 29--November 3, 2023, Ottawa, ON, Canada}
\acmPrice{15.00}
\acmISBN{979-8-4007-0108-5/23/10}
\acmDOI{10.1145/3581783.3612214}

\begin{document}

\title{Whether you can locate or not? Interactive Referring Expression Generation}

\author{Fulong Ye}
\affiliation{%
  \institution{Beijing University of Posts and Telecommunications}
  \city{Beijing}
  \country{China}
  }
\email{fulong_ye@bupt.edu.cn}

\author{Yuxing Long}
\affiliation{%
  \institution{Beijing University of Posts and Telecommunications}
  \city{Beijing}
  \country{China}
  }
\email{longyuxing@bupt.edu.cn}

\author{Fangxiang Feng}
\authornote{Corresponding Author.}
\affiliation{%
  \institution{Beijing University of Posts and Telecommunications}
  \city{Beijing}
  \country{China}
  }
\email{fxfeng@bupt.edu.cn}

\author{Xiaojie Wang}
\affiliation{%
  \institution{Beijing University of Posts and Telecommunications}
  \city{Beijing}
  \country{China}
  }
\email{xjwang@bupt.edu.cn}



\begin{abstract}
    Referring Expression Generation (REG) aims to generate unambiguous Referring Expressions (REs) for objects in a visual scene, with a dual task of Referring Expression Comprehension (REC) to locate the referred object. Existing methods construct REG models independently by using only the REs as ground truth for model training, without considering the potential interaction between REG and REC models. In this paper, we propose an Interactive REG (IREG) model that can interact with a real REC model, utilizing signals indicating whether the object is located and the visual region located by the REC model to gradually modify REs. Our experimental results on three RE benchmark datasets, RefCOCO, RefCOCO+, and RefCOCOg show that IREG outperforms previous state-of-the-art methods on popular evaluation metrics. Furthermore, a human evaluation shows that IREG generates better REs with the capability of interaction \footnote{Ours codes could be found in https://github.com/superhero-7/IREG}.
\end{abstract}




\maketitle

\section{Introduction}

Referring Expressions (REs) are unambiguous language descriptions of objects in visual scenes. They are the bridges connecting language and the physical world, where Referring Expression Generation (REG) aims to generate unambiguous referring expressions for target objects, and Referring Expression Comprehension (REC) is to locate target objects referred by REs. They are essential for many multimodal tasks, and have attracted significant attention from computer vision (CV) and natural language processing (NLP) communities.


Mao et al.\cite{Mao_2016_CVPR} proposed the first end-to-end neural REG model for natural visual scenes and used Maximum Mutual Information~(MMI) training loss to enhance the non-ambiguity of generated RE. Subsequent work has mainly focused on extracting more helpful information from the targets~\cite{yu2016modeling, liu2017referring,liu2020attribute} and their contexts~\cite{li2018bundled,tanaka2019generating,kim2020conan}. These work has greatly contributed to the development of REG. Nevertheless, they all built REG models independently by only using the REs as ground truth to train the model, without considering that a good RE could be used for locating the referred object correctly by a REC model (i.e., REs generated by the REG model can be fed to a REC model.).  The REC model evaluates the REs, and returns information to the REG model for improving the RE. It forms an interaction between the REG and the REC agent. 

\begin{figure}[htbp]
    \centering
    \includegraphics[width=1.0\linewidth]{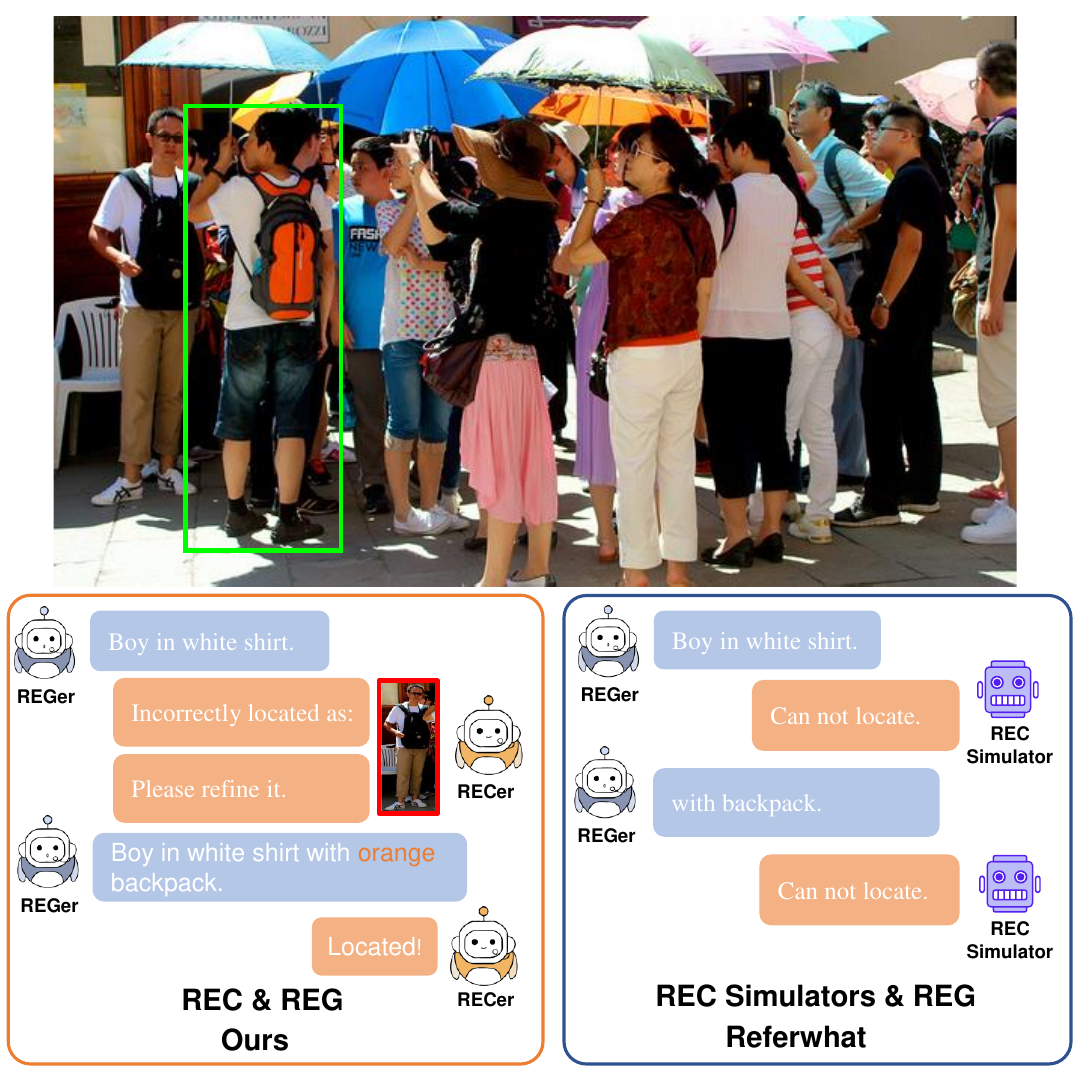}
    \caption{REC model vs. REC simulator in  Referwhat\cite{li2020referring}. The target object is in green bounding box. A real REC model returns additional visual feedback(red bounding box), helps the REG model to add more descriptions, such as "orange". }
    \label{fig:intro}
\end{figure}

Recently, some work jointly trained REG and REC models with shared parameters~\cite{yu2017joint, sun2022proposal, zheng2022towards} for promoting each other. However, essentially, there are no interactions between the REC and REG models. The REG model generates REs without any feedback from the REC model. Li et al.~\cite{li2020referring} proposed a dialogue based REG model Referwhat. The REG model can utilize the feedback from a REC simulator to add more descriptions for referred objects. But the REC simulator in their work is built by REs rather than a real REC model using both REs and visual information. The REC simulator can only tell the REG model if the generated RE is the same as the ground truth, while a REC model can provide more information. For example, in Figure~\ref{fig:intro}, the REG model first generates ``Boy in white shirt''. After receiving the RE generated by the REG model, the REC simulator can only respond with a signal about whether the object can be located. In contrast, the REC model could return not only the signal, but also a region located by the RE, which could be used by the REG model to add new descriptions through comparing the differences between the located region and target objects, such as "orange" in this example, while the REC simulator in Referwhat~\cite{li2020referring} can not do this.  


This paper therefore proposes an Interactive REG~(IREG) model which can explicitly interact with a real REC model, and gradually modify REs by utilizing both the signals of whether the object is located and the visual region located by the REC model. A state-of-the-art (SOTA) REC model OFA~\cite{wang2022ofa} is used as the REC model in the paper. The IREG model is initialized with multimodal pre-trained model VLT5~\cite{cho2021unifying} parameters. A three-stages process is employed to train the IREG. The first stage is fine-tuning for the REG task by using existing RE datasets. It is supervised learning using REs as supervised signals without any interaction with the REC model. The second stage is reinforcement learning. It helps IREG make use of the visual information from the REC model by using the output Intersection-over-Union~(IoU) of the model as a part of the rewards. In the third stage, the IREG is trained to update the RE when the REC model fails to locate the correct object. For this goal, we propose a new refine training task and a dataset for the task. With the capability of updating the RE with the feedback from the REC model, our IREG can conduct multi-round interaction inference. Experimental results on three RE benchmark datasets, RefCOCO, RefCOCO+~(Yu et al.\cite{yu2016modeling}), and RefCOCOg~(Mao et al.\cite{Mao_2016_CVPR}) show our IREG model significantly outperforms previous SOTA methods on popular evaluation metrics. A human evaluation further shows that IREG generates better REs with the ability of interaction.

To summarize, our main contributions are as follows: (1) We propose an IREG model. It is the first REG model that can explicitly interact with a REC model, make use of both the signal and visual information from the REC model. (2) We propose a three-stages process to train our IREG model which can update REs by utilizing feedback from the REC model. (3) Experimental results on RefCOCO, RefCOCO+ and RefCOCOg show that our approach outperforms previous SOTA methods by a large margin.

\section{Related Work}

\subsection{Referring Expression}

Referring expression has been studied in linguistics and natural language processing for many years. There is a pair of dual tasks on referring expression, referring expression comprehension~(REC) and referring expression generation~(REG).

Previous methods of REC can be divided into two types, including two-stage methods and one-stage methods. Given an image and a referring expression, Two-stages methods ~\cite{mattnet,refnms,twostagerec1,twostagerec2} first generate a set of proposal candidate regions, which are usually extracted by a pre-trained object detector. In the second stage, the models calculate the matching score between candidate regions and referring expression and then select the region with the highest matching score. Compared with two-stage approaches, one-stage methods\cite{faoa,resc,rlrec} can reach the goal of real-time processing and achieve superior grounding accuracy. Recently, multi-modal pretrained models~\cite{cho2021unifying, wang2022ofa} achieved pretty well results in one-stage way. 

Early REG work was mainly done on small datasets. In 2014, Kazemzadeh et al.\cite{kazemzadeh2014referitgame} introduced the first large-scale dataset RefCLEF. The authors further collected RefCOCO and RefCOCO+ datasets on MSCOCO images\cite{yu2016modeling}. In addition, Mao et al.\cite{Mao_2016_CVPR} also collected RefCOCOg using MSCOCO images and introduced the first end-to-end deep neural network for REG. After that, some works focus\cite{yu2016modeling, kim2020conan, tanaka2019generating, boc} on extracting more useful visual context to generate unambiguous RE. Liu et al.\cite{liu2020attribute} designed an additional module to incorporate attribution into the generation model.

There are few methods that perform both generation and comprehension tasks. Yu et al.\cite{yu2017joint} first proposed a joint framework to train REG and REC together. Recently, some works have focused on joint learning REG and REC with parameter sharing\cite{sun2022proposal, zheng2022towards}. However, these works still fail to build an explicit interactive framework. Li et al.\cite{li2020referring} first tried to make REG incrementally generate RE by explicitly interacting with a rule-based REC simulator under the dialog framework. In contrast, we first propose an explicit interactive framework including Interactive REG and real REC.

\subsection{Vision-Language Pre-training}

To improve models' perception of text and image and help them establish connections between multimodal information, kinds of visual language pretraining models are designed. ViLBERT~\cite{vilbert} and UNITER~\cite{uniter} propose to consider the raw output of the detector, a distribution of object classes, as soft labels and optimize the KL-divergence between two distributions. LEXMERT~\cite{lxmert} and UNIMO~\cite{unimo} propose Masked Region Feature Regression (MRFR) regresses the masked region feature to its corresponding original region feature, where represents images as a sequence of region features by Faster R-CNN~\cite{ren2015faster}. Furthermore, SOHO~\cite{soho} is designed to avoid information leakage from neighbor features when images are converted into grid features or patch features.

Recently, CLIP~\cite{clip} and ALIGN~\cite{align} leverage large-scale image-text pairs to learn transferable visual representations and exhibit surprising zero-shot transfer to image classification tasks. VLT5~\cite{cho2021unifying} and OFA~\cite{wang2022ofa} introduce downstream tasks, like visual grounding and grounded caption, into pretraining tasks to narrow the gap between pretraining and fine-tuning. Considering good performance of these two vision-language pretrained models, we build REG based on VLT5~\cite{cho2021unifying} and utilize OFA~\cite{wang2022ofa} as REC in our framework.





\section{Proposed Method}
\begin{figure}[htbp]
    \centering
    \includegraphics[width=1\linewidth]{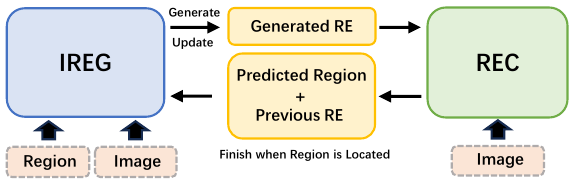}
    \caption{Multi-round Interaction Framework}
    \label{fig:framework}
\end{figure}

As shown in Figure~\ref{fig:framework}, We propose a REG-REC interaction framework including a REC agent and an Interactive REG (IREG) model.

\textbf{\textit{REC Agent.}} 
Unlike the previous method (e.g., Referwhat\cite{li2020referring}), we use a REC model instead of a REC simulator in our interaction framework. The REC agent takes the RE generated by the IREG model as the input and outputs prediction region bounding box, which is more useful feedback information. The prediction region bounding box will be used in section~\ref{rl_training} to calculate reward, in section~\ref{refiner} to help collect Interactive History RE dataset and in section~\ref{interact_infer} as visual feedback to conduct multi-round interactive inference.

\textbf{\textit{Interactive REG.}} 
In our interaction framework, the IREG model can continuously update RE conditioning on the feedback of REC and previous generated RE. We first introduce our REG model architecture in section~\ref{base_model}. Then, we propose a three-stages training strategy to train our IREG. First, we conduct supervised training by using Maximum Mutual Information~~(MMI)\cite{Mao_2016_CVPR} loss in section~\ref{supervised_training}. Then, we do reinforcement training by our proposed novel reward in section~\ref{rl_training}. Finally, we train our IREG based on Reinforced REG in section~\ref{refiner}.

\subsection{Model Architecture\label{base_model}}
Due to the emergency of the multi-modal pretrained model, we can utilize the basic knowledge learned by multi-modal pretrained model to help downstream tasks. Therefore, in this paper, We construct our REC and REG model based on the multi-modal pretrained model.

\subsubsection{\textbf{REC Model Architecture}} \ 

\noindent The backbone model of REC is OFA \cite{wang2022ofa}, which is an encoder-decoder transformer based single-stream vision language pre-trained model. We directly adopt it to conduct REC task by using task instruction ``visual grounding''. We suggest readers to check OFA~\cite{wang2022ofa} for more details of our backbone models. In this paper, we focus more on building an IREG model to interact with the REC model, so we only use REC for inference and freeze its parameter. We call the OFA REC as RECer in the rest part of this paper.
\subsubsection{\textbf{REG Model Architecture.\label{reg_model_architecture}}} \ 

\noindent The backbone model of REG is a vision and language transformer architecture with a bidirectional multimodal encoder and auto-regressive text decoder. The bidirectional multimodal encoder is a stack of transformer blocks consisting of a self-attention layer and a fully-connected layer with residual connections. The auto-regressive text decoder is another stack of transformer blocks similar to the multimodal encoder, where each block has an additional cross-attention layer.

\begin{figure*}[ht] 
\centering 
\includegraphics[width=0.90\textwidth]{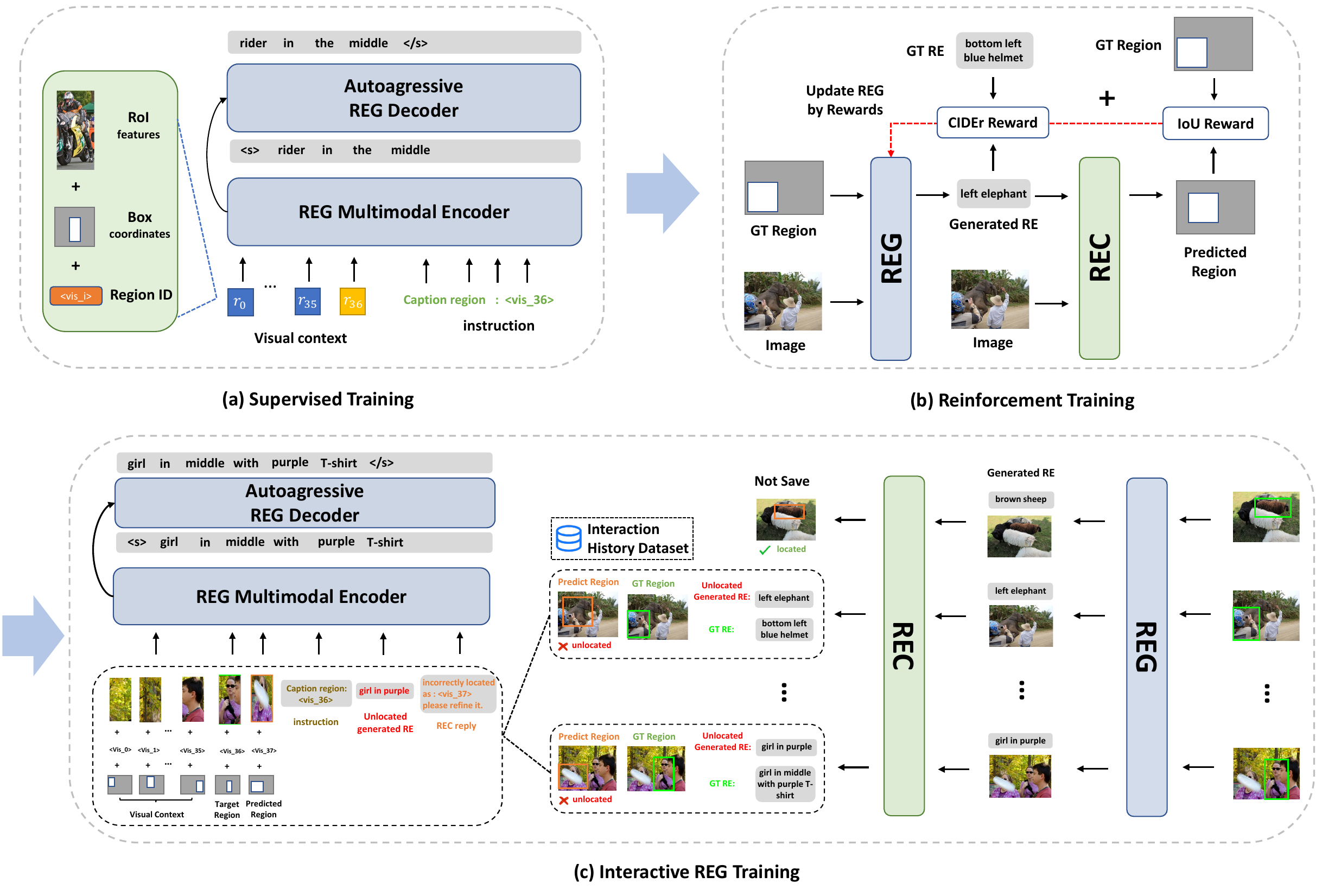} 
\caption{An illustration of Three-stage training strategy: (a)-->(b)-->(c). (a) illustrates the REG task. (b) is reinforcement training using CIDEr and Rec reward. Data collection in the right part of (c) mimics REG and REC interaction to gather Interaction History Data. The left part of (c) illustrates the Refiner task. All REG models share the same architecture.}
\label{fig:training}
\end{figure*}

\textbf{\textit{Visual Embedding.}} 
We represent an input image $I$ by $n=36$ object regions extracted from Faster R-CNN\cite{ren2015faster}. As shown in Figure~\ref{fig:training}(a), each object region feature $e_{i}^{v}(i\in[0,35])$ is the sum of: (i) RoI~(Region of Interest) object features; (ii) RoI bounding box coordinates; (iii) Region ID which is textual embedding of visual sentinel token. The target object region $R$ is encoded to $e_{36}^{v}$ and put in the last position. We denote the final visual embedding as $e^{v} = \{e_0^{v},...,e_{35}^{v}\} + \{e_{36}^{v}\}$.

\textbf{\textit{Textual Embedding.}} 
Following the recent popular technique\cite{wang2022ofa,cho2021unifying}, We add a instruction "\textit{caption region: $<vis\_36>$}" to specify the REG task and target object, which is then tokenized to $p=\{p_1,...,p_{|p|}\}$ and encoded as learned embedding $e^{p} = \{e_1^{p}, ..., e_{|p|}^{p}\}$. In particular, the visual sentinel tokens $\{<vis_1>, ... , <vis_n>\}$ will be added to the vocabulary as a special token, and will not be split during tokenization, in order to be used to indicate a specific object. The region id in object region visual embedding is represented by textual embedding of visual sentinel tokens.

As shown in Figure~\ref{fig:training}(a), the encoder takes the concatenation of textual embedding and visual embedding as input and outputs their contextualized joint presentation:
\begin{equation}
h = \{h_1^{p},...,h_{|p|}^p,h_1^v,\\ ...,h_{37}^v\}=Enc(e^p,e^v)    
\end{equation}
 Then the decoder predicts the probability of future text tokens $P_\theta(y_j|y_{\textless j},p,v)=Dec(y_{\textless j},h)$ in an auto-regressive way.


\subsection{Model Training \label{training}}
The model is trained in a three-stages process. The first stage is supervised training, REs are used as supervised signals without interaction with the REC model. Existing RE datasets are used for the first stage training. The second stage is reinforcement learning. Visual information (Intersection-over-Union(IoU)) from the REC model is used as a part of the reward signal. The third stage is for building the capability of interaction between the IREG model and the REC model. The IREG model is trained to update the RE when it receives both the RE generated in last round and the visual location information from the REC model. A dataset for the training is also built. We give the details of the three-stages one by one.
\subsubsection{\textbf{Supervised Training.} \label{supervised_training}} \ 

\noindent Base on the model architecture described in section~\ref{reg_model_architecture}, we conduct supervised training by the RE datasets. Our training data consists of observed triplets $(I,R,T)$, where $I$ is an image, $R$ represents an object region in $I$, and $T$ denotes a referring expression for $R$. Then we minimize the negative log probability of the referring expressions given the region and image:
\begin{equation}\label{base_loss}
J(\theta)=-\sum_{n=1}^N \log p\left(T_n \mid R_n, I_n, \theta\right)
\end{equation}
\noindent where $\theta$ represents model parameters of the Transformer encoder and decoder, and $N$ is the number of data samples in the training set. Furthermore, we follow Mao et al.\cite{Mao_2016_CVPR} to calculate MMI loss to train our model:
\begin{equation}\label{mmi_loss}
\begin{aligned}
& J^{s}_{1}(\theta)=-\sum_{n=1}^N\left\{\log p\left(S_n \mid R_n, I_n, \theta\right)-\right. \\
& \left.\quad \lambda \max \left(0, M-\log p\left(S_n \mid R_n, I_n, \theta\right)+\log p\left(S_n \mid R_n^{\prime}, I_n, \theta\right)\right)\right\}
\end{aligned}
\end{equation}
\noindent where $R^{'}$ is another region in $I$ with the same class of $R$. In subsequent sections, we build the following REG model from the MMI REG base model.



\subsubsection{\textbf{Reinforcement Training.} \label{rl_training}} \ 

\noindent Previous works\cite{yu2017joint, tanaka2019generating, li2020referring} adopt CIDEr reward to conduct reinforcement learning, which can significantly improve the model performance on automatic metrics. To mitigate the challenges associated with interactive learning, we design a novel reward that leverages RECer to incentivize the REG model to generate REs that are more easily locatable.

As shown in Figure~~\ref{fig:training}(b), given $\{I,R,T\}$, we use the MMI REG base model to generate RE $T_{generate}$. Then we input $I$ and $T_{generate}$ into RECer to get predicted $R_{predict}$, calculating the IoU between $R$ and $R_{predict}$ as reward. The formulation is shown in the following:
\begin{equation}
    Rec\_reward = IoU(BBox(R),\,BBox(R_{predict}))
\end{equation}
\noindent where $BBox(\cdot)$ means the bounding box of object region. We note that only using Rec\_reward will suffer heavy language drift in section~\ref{alation_study_section}, so we combine the CIDEr reward and Rec reward as our final reward. The formulation is shown bellow:
\begin{equation}
    Reward = Rec\_reward + \beta\,CIDEr(T,T_{generate})
\end{equation}
\noindent where $CIDEr(\cdot)$ means calculating the CIDEr score between $T$ and $T_{generate}$. Finally, we use this reward to train our Reinforced REG model by REINFORCE algorithm\cite{williams1992simple}. The training loss is shown bellow:
\begin{equation}
\nabla_\theta J^{r}=E_{p\left(T\right)}\left[Reward \cdot \nabla_\theta \log p\left(T ; \theta\right)\right]
\end{equation}
During the reinforcement learning training process, the newly proposed REC reward incentivizes the REG to generate REs that are easily locatable, while the CIDEr reward maintains the fluency of the language. After reinforcement learning training, the resulting model is referred to as the Reinforced REG model, which is capable of generating REs that are both fluent and unambiguous. This model serves as a solid foundation for training IREG. Therefore, we further train the IREG model based on the Reinforced REG model and fully use the feedback from REC.

\subsubsection{\textbf{Interactive REG Training}\label{refiner}} \ 

\noindent In order to make REG have the ability to update REs using previous imperfect generated REs and visual information from REC feedback. As shown in Figure~\ref{fig:training}(c), We design a new refiner training task. In addition, we collect a new Interactive History RE dataset to support the refiner training task. 


\textbf{\textit{Refiner Training Task.}}
As shown in the left part of Figure~\ref{fig:training}(c), We extend the instruction from only ``\textit{caption region: $<vis\_36>$}'' in supervised learning to ``\textit{caption region: $<vis\_36>$ + $T_{generate}$ + incorrectly located as: $<vis\_37>$ + Please refine it.}''. We also add the region feature of $R_{predict}$ into $e^{v}$. That means $e^{v} = \{e_0^{v},...,e_{35}^{v}\} + \{e_{36}^{v}\} + \{e_{37}^{v}\}$, $e_{37}^{v}$ is the region feature of $R_{predict}$. The training loss can be formalized as follows:
\begin{equation}
J^s_{2}(\theta)=-\sum_{i=1}^M \log p\left(T_i \mid R_i, I_i, R_{predict}^i, T_{generate}^i,\theta\right)
\label{refiner_loss}
\end{equation}
\noindent We refer to Equation~~\ref{refiner_loss} as refiner loss, which encourages REG model to generate a better RE conditioning on generated RE $T_{generated}$ and $R_{predict}$. This means that model can learn to refine a bad generated RE by using refiner loss. However, if only this loss is used, the model's original ability to generate REs will be compromised. To address this issue, we employed the Round Robin training strategy for multitask training\cite{cho2021unifying}, which involves simultaneously training the REG and refiner task. After interactive training, we obtain an IREG model that can generate and modify REs. As a result, we can utilize IREG to do multi-rounds interactive inference in section~\ref{interact_infer}.

\textbf {\textit{Interaction History Data Collection.}}
To support the refiner training task, we mimic the interactive process between REG and REC to collect Interaction History Data, which is illustrated in the right part of Figure~\ref{fig:training}(c). Given image $I$ and its corresponding target region $R$, we employ our Reinforced REG model to generate RE $T_{generate}$. Then, we use RECer to obtain the predicted region $R_{predict}$, followed by calculating the IoU between $R$ and $R_{predict}$. If the IoU exceeds 0.5, the localization is considered successful, otherwise it fails. We collect the failed locating data $T_{generate}$ and $R_{predict}$, and combine this data with Ground Truth RE dataset $\{I^i, R^i, T^i\}$. By doing so, we finally obtain the Interaction History RE dataset $\{I^i, R^i, T^i, R_{predict}^i, T_{generate}^i\}$, where $i \in M$, $M$ is the total number of collected data. 

\begin{algorithm}[tb]
    \caption{Multi-Rounds Interactive Inference Algorithm}
    \label{alg:algorithm}
    \renewcommand{\algorithmicrequire}{\textbf{Input:}}
    \renewcommand{\algorithmicensure}{\textbf{Output:}}
    \begin{algorithmic}[1] 
        \REQUIRE $I$,$R$
        \ENSURE Referring Expression $T$
        \STATE Init $T_{0}^{'}$ $=$ Reinforced REG($I$, $R$)
        \STATE Let $MAX\_ROUND=n$, $Cur\_ROUND \ t=0$.
        \WHILE{$0 \textless t \textless n$}
        \STATE Get the predict region:\\
        $R_{predict}^{t-1}$ $=$ RECer($I$, $T_{t-1}^{'}$)
        \STATE Calculating the IoU between $R$ and $R_{predict}^{t-1}$:\\
        $Score=IoU(R, R_{predict})$
        \IF {$Score \textgreater 0.5$}
        \STATE \textbf{return} $T^{'}_{t-1}$
        \ELSE
        \STATE Refine RE: \\ 
        $T^{'}_{t}$=IREG($I,R,R_{predict}^{t-1},T^{'}_{t-1}$)
        \ENDIF
        \ENDWHILE
        \STATE \textbf{return} $T_{n-1}^{'}$
    \end{algorithmic}
\end{algorithm}

\subsection{Multi-round Interaction Inference\label{interact_infer}}

Once the three training stages are completed, the IREG model can be utilized to engage in multi-round interactions with the REC model.

To illustrate the inference process more clearly, we show it in Algorithm~\ref{alg:algorithm}. Let $MAX\_ROUND=n$ and give image I and region R. When $n=0$, we generate initial RE by the following equation:
\begin{equation}
T_{0}^{'}=argmax \,  P_{\theta_{Reinforced \ REG}} \left(T_{0} \mid I, R\right)
\end{equation}
where the $\theta_{IREG} $ means the parameter of Reinforced REG model. When the current round $t$ less than $n$, we utilize RECer to abtain predicted region $R_{predict}$. We then caculate IoU between $R$ and $R_{predict}$. If the IoU is greater than 0.5, $T^{'}_{t-1}$ is returned. Otherwise, IREG is utilized to refine the generated RE in the First Order Markov Chain way like the following equation:
\begin{align}
T_{t}^{'} &=argmax \,  P_{\theta_{IREG}} \left(T_{t-1} \mid I, R, R_{predict}^{t-1}, T_{t-1}^{'}, ... ,T_{0}^{'}\right) \\
&=argmax \,  P_{\theta_{IREG}} \left(T_{t-1} \mid I, R, R_{predict}^{t-1}, T_{t-1}^{'}\right)
\end{align}
If no successful localization is achieved by RECer at time step n, the result from the final iteration $T_{n-1}^{'}$ is returned.


\section{Experiments}
\subsection{Datasets and Metrics\label{datasets_and_metrics}}


\begin{table*}[]
\centering
\caption{Experimental results of different models on three data sets. Our IREG model significantly outperforms all existing models on all metrics.(The results of the second-best model are marked with underline.)}
\renewcommand\arraystretch{1.0}
\resizebox{1\textwidth}{!}{
\setlength{\tabcolsep}{3mm}{
\begin{tabular}{l|cccc|cccc|cc} \hline
\multirow{3}{*}{Method} & \multicolumn{4}{c}{RefCOCO} & \multicolumn{4}{|c|}{RefCOCO+} & \multicolumn{2}{c}{RefCOCOg} \\ \cline{2-11}
& \multicolumn{2}{c}{testA} & \multicolumn{2}{c}{testB} & \multicolumn{2}{|c}{testA} & \multicolumn{2}{c|}{testB} & \multicolumn{2}{c}{val} \\ \cline{2-11}
& Meteor & CIDEr & Meteor & CIDEr & Meteor & CIDEr & Meteor & CIDEr & Meteor & CIDEr \\ \hline
Visdif\cite{yu2016modeling} & 0.185 & - & 0.247 & - & 0.142 & - & 0.135 & - & 0.145 & -\\
SLR\cite{yu2017joint} & 0.296 & 0.775 & 0.340 & 1.320 & 0.213 & 0.520 & 0.215 & 0.735 & 0.159 & 0.662 \\
Attr\cite{liu2020attribute} & 0.312 & 0.802 & 0.332 & 1.301 & 0.236 & 0.585 & 0.206 & 0.692 & 0.163 & 0.645 \\
PFOS\cite{sun2022proposal} & 0.303 & 0.877 & 0.330 & 1.333 & 0.253 & 0.722 & 0.210 & 0.758 & 0.156 & 0.749 \\
EU\cite{tanaka2019generating} & 0.313 & 0.837 & 0.341 & 1.329 & 0.242 & 0.664 & 0.228 & 0.787 & 0.170 & 0.777 \\
CoNAN\cite{kim2020conan} & \underline{0.330} & \underline{0.915} & 0.354 & 1.410 & \underline{0.288} & \underline{0.761} & \underline{0.250} & 0.876 & \underline{0.183} & \underline{0.910} \\
ReferWhat\cite{li2020referring} & 0.326 & 0.914 & \underline{0.366} & \underline{1.473} & 0.258 & 0.684 & 0.247 & \underline{0.895} & - & - \\ \hline
IREG & \textbf{0.349} & \textbf{1.054} & \textbf{0.373} & \textbf{1.541} & \textbf{0.308} & \textbf{0.898} & \textbf{0.264} & \textbf{0.970} & \textbf{0.194} & \textbf{1.012} \\
\hline
\end{tabular}}
}
\label{table_performace_reg}
\end{table*}

\begin{table}[]
\centering
\caption{Ablation study on different combinations of all rewards and interactive training.}
\renewcommand\arraystretch{1.4}
\resizebox{\columnwidth}{!}{%
\setlength{\tabcolsep}{1mm}{
\begin{tabular}{ll|cccccc} \hline
\multirow{3}{*}{} & \multirow{3}{*}{Method} & \multicolumn{6}{c}{RefCOCO+} \\ \cline{3-8}
& & \multicolumn{3}{c}{testA} & \multicolumn{3}{c}{testB} \\ \cline{3-8}
& & Meteor & CIDEr & Acc & Meteor & CIDEr & Acc \\ \hline
& MMI & 0.287 & 0.838 & 0.768 & 0.241 & 0.857 & 0.558 \\
\hline
\multirow{3}{*}{\textbf{Reward}} 
& \quad w/ Rec & 0.214 & 0.552 & \textbf{0.802} & 0.221 & 0.745 & \textbf{0.613} \\
& \quad w/ CIDEr & 0.299 & 0.877 & 0.763 & 0.253 & 0.948 & 0.552 \\
& \quad w/ CIDEr+Rec & \textbf{0.302} & \textbf{0.881} & 0.782 & \textbf{0.259} & \textbf{0.955} & 0.586 \\
\hline 
\textbf{Refiner} & \quad w/ Refiner & 0.290 & 0.846 & 0.839 & 0.243 & 0.861 & 0.668 \\
\hline
\textbf{All} & \quad w/ CIDEr+Rec+Refiner & \textbf{0.308} & \textbf{0.898} & \textbf{0.851} & \textbf{0.264} & \textbf{0.970} & \textbf{0.675} \\
\hline
\end{tabular}}
}
\label{abalation}
\end{table}

\textbf{\textit{Datasets.}} We conduct experiments on three widely-used benchmark datasets: RefCOCO, RefCOCO+~\cite{yu2016modeling} and RefCOCOg~\cite{Mao_2016_CVPR}, which are all collected on MSCOCO images\cite{lin2014microsoft}. RefCOCO contains 142,209 reference expressions for 50,000 objects on 19,994 images, and RefCOCO+ consists of 141,564 descriptions for 50,000 objects on 19,992 images, while ReCOCOg contains 54,822 objects on 26,711 images with 104,560 expressions. RefCOCO and RefCOCO+ were collected in an interactive setting, where annotators aimed to complete tasks quickly, resulting in concise REs. In addition, RefCOCO+ prohibited the use of directional words during data collection, making it more focused on appearance-based descriptions, e.g., "the man in the yellow polka-dotted shirt" rather than "the second man from the left" ,and thus more challenging. In contrast, RefCOCOg was collected in a non-interactive setting, resulting in REs that are more elaborate.

\textbf{\textit{Interaction History Data.}} As described in section ~\ref{refiner}, we mimic the interactive process between REG and REC to collect Interaction History Data. We utilize the RECer to filter the REs generated by Reinforced REG, and combine the unsuccessful located REs with original dataset to create a new dataset. The proportion of unsuccessfully located REs in the RefCOCO, RefCOCO+, and RefCOCOg datasets are 22\%, 46\%, and 39\%, respectively. This distribution aligns with the difficulty of the datasets, where RefCOCO+ is the most challenging due to the limitation on directional words, followed by RefCOCOg, which has longer sentences, while RefCOCO is the easiest.

\textbf{\textit{Metrics.}} We adopt two widely-used automatic metrics, CIDEr and Meteor, to evaluate the performance of REG. In order to further evaluate our method, we propose a new metric named REC Accuracy. Specifically, we get the predicted region from RECer $R_{predict}$ by taking the generated RE $T_{generate}$ as input. Then, we compute the IoU of the $R_{predict}$ and ground truth object $R$. If the IoU score exceeds 0.5, we consider the generated RE can be located correctly. Dividing the number of samples with correctly located RE by the total number of samples can obtain the REC Accuracy score.

\subsection{Implementation Details}



We initialized the RECer with the OFA-base checkpoint, which contains 180M parameters, and froze the parameters of OFA. The REG model was initialized with VLT5 pretrained checkpoint, which contains 220M parameters. We implemented our method using Pytorch and conducted all experiments on 4 NVIDIA A40 GPUs with 48G memory. In the supervised learning, we trained our REG base model starting from the VLT5 pre-trained checkpoint with a learning rate of 5e-5 and a warm-up ratio of 0.1 for 30 epochs. For reinforcement learning, the Reinforced REG model was fine-tuned from the REG base model with a learning rate of 5e-06 and a warm-up ratio for 5 epochs. We used the same learning rate of 5e-06 in the IREG model to train for 20 epochs based on Reinforced REG. During inference, we used a beam size of 5 for beam search decoding. The $\beta$ in section~\ref{rl_training} is set to 0.5, and $MAX\_ROUND$ in section~\ref{interact_infer} is 5.

\subsection{Main Results}
In this section, we compare our IREG model with several baselines and SOTA models of REG.
\textbf{(1)} Visdif~\cite{yu2016modeling} incorporates better visual context by utilizing visual difference features into referring model.
\textbf{(2)} SLR~\cite{yu2017joint} proposes the speaker-listener-reinforcer framework.
\textbf{(3)} Attr~\cite{liu2020attribute} designs an additional attribute module and incorporates it into the generation model.
\textbf{(4)} PFOS~\cite{sun2022proposal} proposes an REG-REC joint training framework by parameter sharing.
\textbf{(5)} EU\cite{tanaka2019generating} extends the SLR with a well-designed attention mechanism. 
\textbf{(6)} CoNAN~\cite{kim2020conan} introduces an attentional ranking module to obtain complementary neighbor features.
\textbf{(7)} Referwhat~\cite{li2020referring} designs a REG model that can generate RE incrementally under the framework of dialog.

Table~\ref{table_performace_reg} shows that our IREG model significantly outperforms all existing models on all metrics. In terms of CIDEr, IREG achieves 1.054/1.541 on RefCOCO testA/testB set, 0.898/0.970 on RefCOCO+ testA/testB set, and 1.012 on RefCOCOg val set. Specifically, IREG exhibits the most significant advancement on RefCOCO+, with CIDEr lift rate of 18 \% and 8.4 \% on testA and testB respectively, compared to the second-best performing model. RefCOCO+ represents the most challenging dataset, as it precludes the use of positional language, meaning that the model must identify the most salient attributes of the object of interest to generate clear and unambiguous referring expressions, instead of relying on positional language (e.g., "in the left, in the middle"). Under such circumstances, interaction with REC can be immensely beneficial, which explains why IREG exhibits the most significant improvement on RefCOCO+.

\subsection{Ablation Studies \label{alation_study_section}}

\begin{figure*}[ht] 
\centering 
\includegraphics[width=1.0\textwidth]{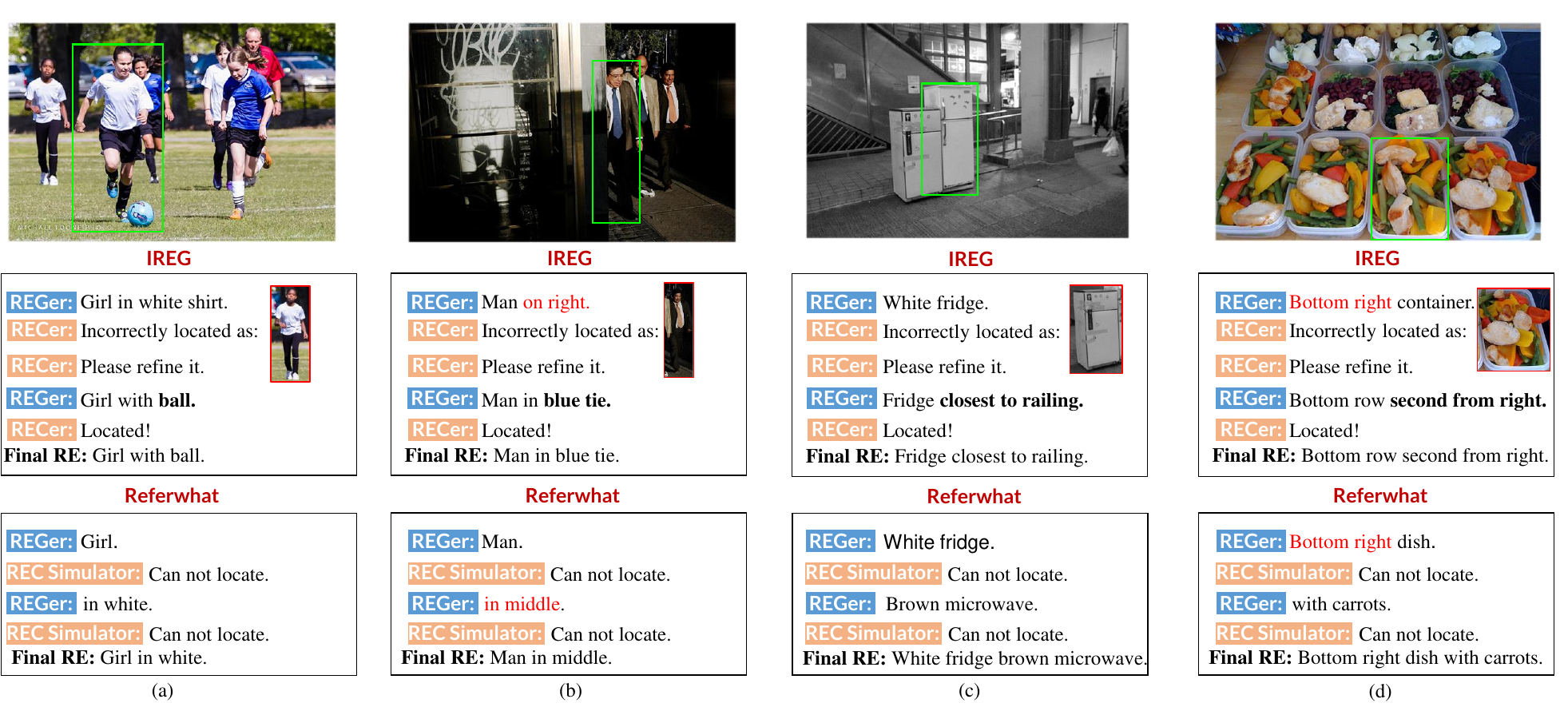} 
\caption{Comparison of the generation results between IREG and Referwhat. The target object is highlighted by a green bounding box. The object highlighted by a red bounding box in IREG is the incorrectly localized object by the RECer. Compared to Referwhat, IREG can utilize visual feedback to supplement key differentiating features or correct generated errors.}
\label{fig: compared_results}
\end{figure*}


Table~\ref{abalation} shows the results of our ablation experiments, where MMI refers to the base REG model trained with MMI loss, Rec and CIDEr refer to two types of rewards, and Refiner refers to the model trained with refiner loss in Equation~\ref{refiner_loss}. We design the ablation experiments from two aspects. First, we study the impact of different rewards on the Reinforced REG model. Second, we validate the effectiveness of interactive training.

\textbf{\textit{Rewards}} Comparing MMI and MMI with Rec reward, we observed a significant improvement in REC Accuracy with the latter, but a decrease in CIDEr. This suggests that using only the Rec reward may result in REs preferred by the REC model but with significant language drift. To address this, we combined the Rec reward, responsible for improving non-ambiguity, with the CIDEr reward to maintain language coherence. The results of the combined reward (MMI with CIDEr + Rec) show improvements in both REC Accuracy and CIDEr compared to MMI.

In addition, we conducted an experiment using only the CIDEr reward (MMI + CIDEr). Comparing it with MMI + CIDEr + Rec, we observed that the Rec reward not only improves REC Accuracy, but also slightly enhances CIDEr. This finding indicates that the CIDEr and Rec rewards complement each other. Our newly proposed REC reward encourages the generation of REs that are easier to locate and closer to ground truth.



\textbf{\textit{Refiner}} We perform interactive training using the Refiner loss (Equation~\ref{refiner_loss}) based on the MMI checkpoint. Comparing the MMI and MMI with Refiner, we can observe that training with Refiner loss significantly improves the REC Accuracy while slightly enhancing CIDEr. This indicates that the REG model trained with Refiner loss can generate REs that are easier to locate through multi-round interactions with REC. 

Based on the Reinforced REG (MMI with CIDEr + Rec), we also perform interactive training to investigate the correlation between reinforcement learning and interactive learning. The results show that IREG (MMI with CIDEr + Rec + Refiner) achieves higher REC Accuracy than Reinforced REG, demonstrating that interactive learning can further improve the generative quality of reinforcement learning.




\subsection{Qualitative analysis}
We compare the generation results of our IREG and Referwhat in the section~\ref{compared_results}. Then, in the section~\ref{case_study}, we present some examples of successful and unsuccessful modifications made by IREG for case study.

\subsubsection{Generated Results compared with Referwhat \label{compared_results}} 

We compared the generated results of IREG with Referwhat which is we re-implemented. For the convenience of display, we only show examples of two rounds of dialogue here. More generated results can be found in the supplementary materials.

As shown in Figure~\ref{fig: compared_results}, REC simulator in Referwhat can only determine the success of the localization based on rule matching, whereas IREG employs a real REC model, which can provide visual feedback of incorrectly localized objects. Therefore, IREG can supplement salient features by comparing the differences between the incorrectly localized object and the target object, such as ``ball'' in (a), ``blue tie'' in (b), ``closest to the railing'' in (c), and ``second from right'' in (d). 

Furthermore, Referwhat can only continually supplement information by breaking down a complete RE into several parts for generation. However, this approach relies on the strong assumption that the previously generated information must be correct. If the previous information is incorrect, Referwhat cannot make any modifications, as shown in (d) where Referwhat generated the incorrect information ``Bottom right dish'' in the first round but continued to add useless information ``with carrots'' in the second round. In contrast, IREG can correct its own generated errors, such as removing ``on the right'' in (b) and changing ``Bottom right'' to ``Bottom row second from right'' in (d).

\begin{figure}[htbp] 
\centering 
\includegraphics[width=1.0\linewidth]{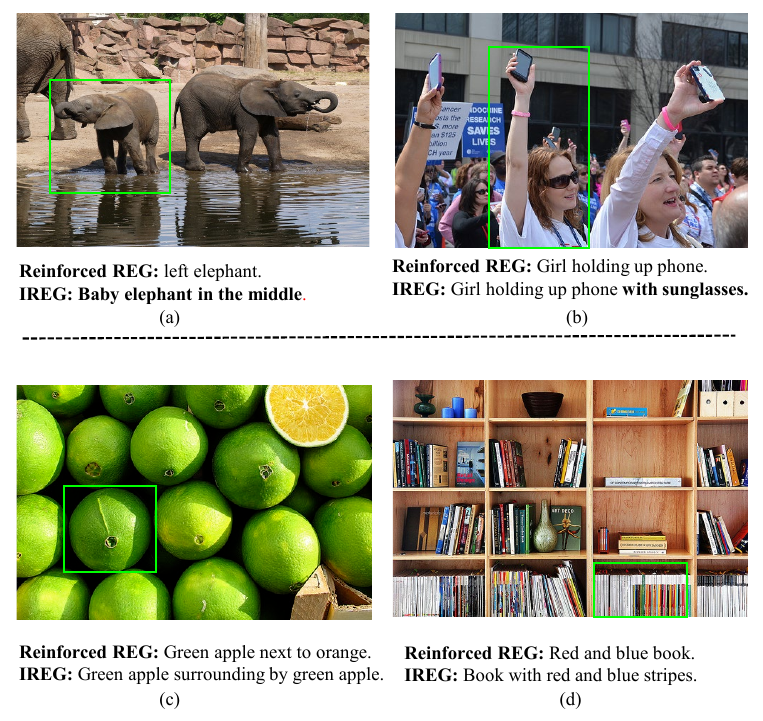} 
\caption{Generated examples of Reinforced REG and IREG. The first line is a correct example modified by IREG, and the second line is an example of failure.}
\label{fig: generated_examples}
\end{figure}

\subsubsection{Case study \label{case_study}} Figure~\ref{fig: generated_examples} shows generated examples of Reinforced REG and IREG. The green bounding box points out the target object. To investigate the role of IREG, we selected examples of REs generated by Reinforced REG in the first round that were not successfully located by the RECer. More generated results of Reinforced REG can be found in the supplementary meterials.

The first line in Figure~\ref{fig: generated_examples} is a correct example modified by IREG, as mentioned in the section~\ref{compared_results}, IREG can modify errors in previous REs (e.g. ``left elephant'' --> ``Baby elephant in the middle'' in (a)) or add crucial distinguishing features(e.g. ``with sunglasses'' in (b)). And the second line is an example of failure modified. Most of the unsuccessful examples come from RefCOCO+, as it limits the use of positional words. In scenes with multiple similar objects(e.g. (c), not using positional words can greatly increase the difficulty of the task. In such cases, IREG will use clock-based expressions like "at 9 o'clock" as a "cheat", but the performance remains unsatisfactory. Since the target object lacks distinctive features(e.g the book in (d)), it is also difficult for humans to identify it without using directional information.

\subsection{Human Evaluation}
To futher evaluate the quality of our generated RE, we follow previous works ~\cite{yu2017joint, yu2016modeling, liu2020attribute, sun2022proposal} to conduct human evaluation: \textit{if a human evaluator can successfully locate the target object on the image by given RE, it is regarded as a correct one. Otherwise, the RE is labeled as an incorrect one.} We randomly sample 100 images for each test split. The human evaluation metric is defined as the proportion of correct REs to total sample REs in each dataset:
\begin{equation}
\frac{\#~correct~RE}{\#~total~sample~REs} 
\end{equation}
 The image with bounding box of the target object and generated RE are displayed to the evaluator to judge whether the target object can be located disambiguously according to the RE. Each test split is distributed to 5 evaluators, and the final result is the average of the 5 evaluators.

We display the human evaluation results in Table~\ref{table:human_eval}. We use the reported results of visdif\cite{yu2016modeling}, SLR\cite{yu2017joint}, Attr\cite{liu2020attribute}, and PFOS\cite{sun2022proposal} in their respective papers, while the results for Referwhat were re-evaluated after our re-implement. The performance of our Reinforced REG has surpassed all previous work. Moreover, our IREG achieve the higher accuracy of human evaluation. Compared with Reinforced REG, IREG exhibits significant improvement on RefCOCO+ and RefCOCOg, which suggests that the interaction process is more advantageous on more challenging dataset.

\begin{table}[t]
\caption{Human evaluation on Referring Expression Generation.}
\scriptsize
\renewcommand\arraystretch{1.7}
\centering
\resizebox{\columnwidth}{!}{%
\begin{tabular}{ l | c | c | c | c | c }
\hline
\multirow{2}{*}{\textbf{Method}} & \multicolumn{2}{c}{RefCOCO} & \multicolumn{2}{c|}{RefCOCO+} & Refcocog\\
\cline{2-6}
&\ \ Test A\ \  &\ \ Test B\ \   &\ \ Test A\ \  &\ Test B\ \ &\ \ val\ \\
\hline\hline
visdif\cite{yu2016modeling} & 70.01\% & 76.31\% & 55.64\% & 48.04\% & - \\
SLR\cite{yu2017joint} & 76.95\% & 78.10\% & 58.85\% & 58.20\% & - \\
Attr\cite{liu2020attribute} & 83\% & 87\% & 49\% & 46\% & - \\
PFOS\cite{sun2022proposal} & 87\% & 84\% & 55\% & 53\% & 61\%\\
Referwhat\cite{li2020referring} & 89.05\% & 86.52\% & 59.62\% & 56.98\% & 65.91\%\\
\hline
Our Reinforced REG & 90.05\% & 88.52\% & 62.62\% & 60.98\% & 71.91\%\\
Our IREG & \bf{92.23\%} & \bf{90.66\%} & \bf{69.33\%} & \bf{65.51\%} & \bf{79.33}\%\\
\hline
\end{tabular}
}
\label{table:human_eval}
\end{table}

\section{Conclusion and future work}

In this paper, we propose the first Interactive REG (IREG) model, which can explicitly interact with a real REC model. Specifically, we design a three-stages training strategy to train our IREG, including supervised training, reinforcement training, and interactive training. Our IREG could update REs gradually by utilizing feedback from the REC. The experiment on RefCOCO, RefCOCO+ and RefCOCOg shows that our IREG significantly outperforms previous SOTA methods. We conduct human evaluation to further evaluate our method. And the improvement in human evaluation is highly consistent with our experiment. 

In the future, we plan to improve REC to a higher level that can ask questions, not just be responsible for locating the target pbject. In this way, REG can supplement information more accurately according to the questions raised by REC. Additionally, we can also leverage ChatGPT to optimize our data collection process. More powerful multimodal pre-training models will also be considered.

\section{Acknowledgements}

The work was supported by the National Natural Science Foundation of China (NSFC62076032). We would like to thank Duo Zheng for the helpful discussions and suggestions. We also thank anonymous reviewers for their suggestions and comments.

\bibliographystyle{acmart}
\balance
\bibliography{acmart}

\clearpage


\begin{figure*}[ht] 
\centering 
\includegraphics[width=1.\textwidth]{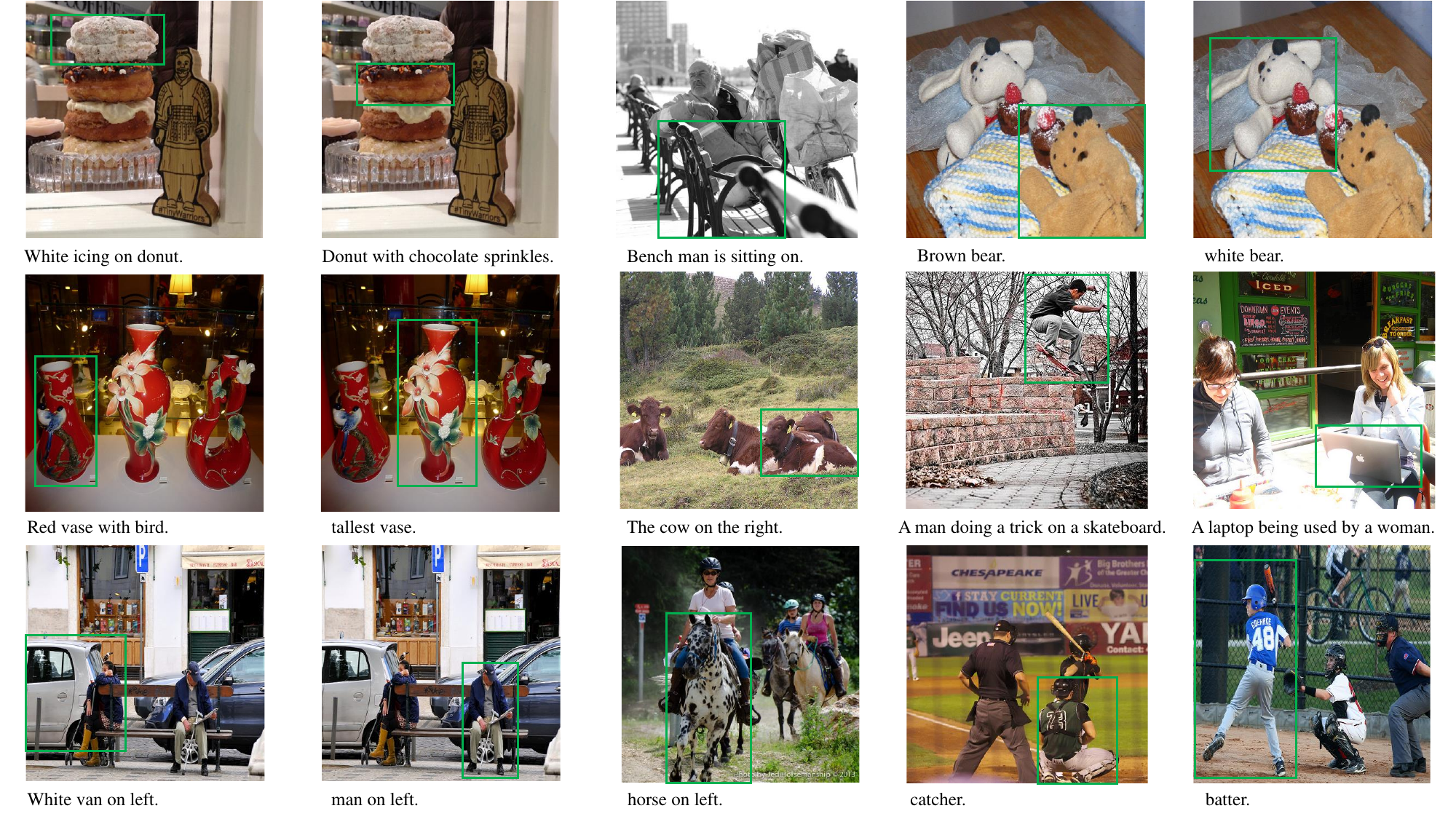} 
\caption{Generated examples of Reinforced REG.}
\label{fig: generated_examples_reinforced_reg}
\end{figure*}

\begin{figure*}[ht] 
\centering 
\includegraphics[width=1.\textwidth]{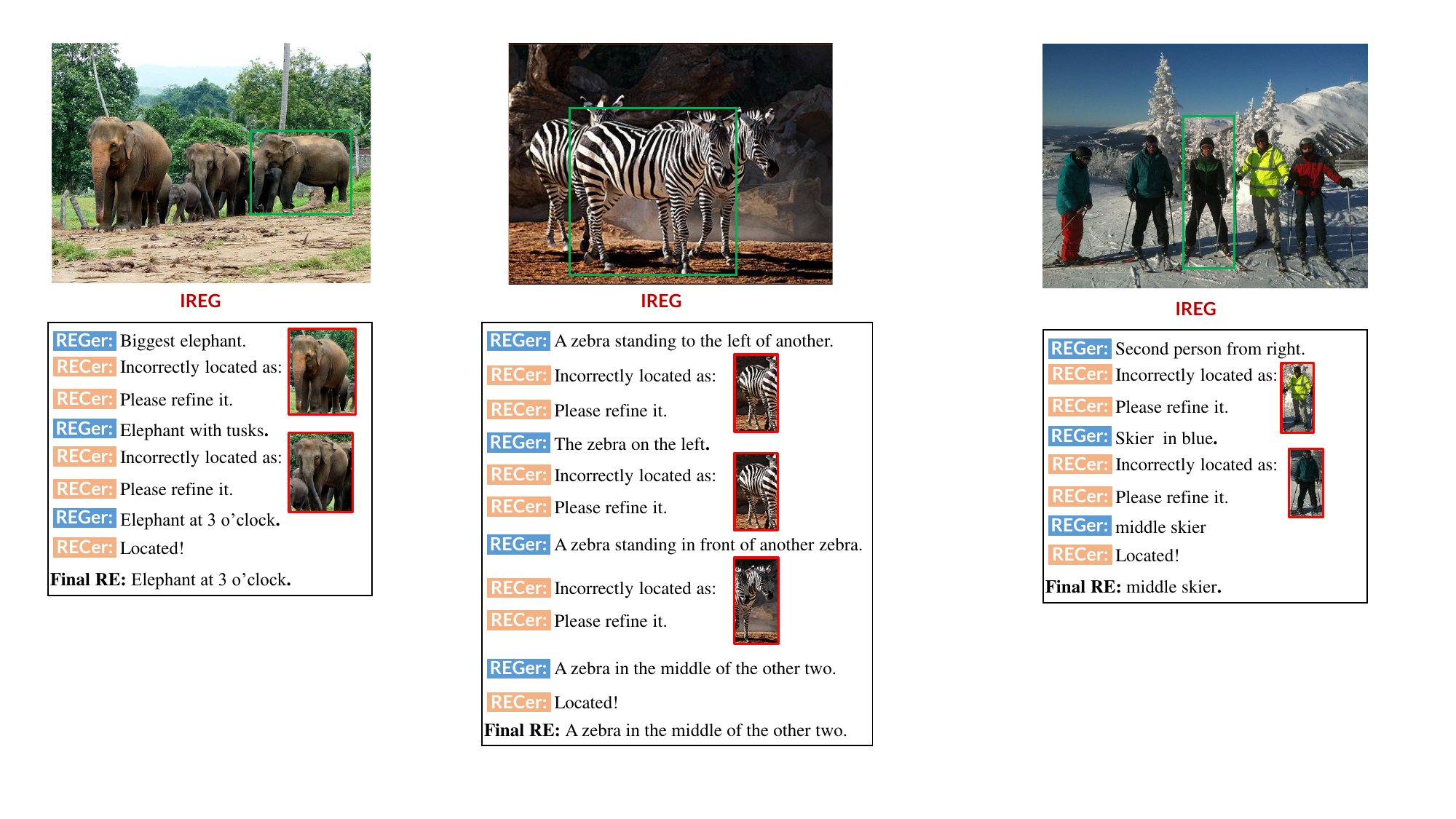} 
\caption{Generated examples of IREG in multi-round.}
\label{fig: generated_examples_IREG_multi_round}
\end{figure*}

\end{document}